\documentclass[10pt,twocolumn,letterpaper]{article}

\usepackage{wacv}
\usepackage{times}
\usepackage{epsfig}
\usepackage{graphicx}
\usepackage{amsmath}
\usepackage{amssymb}
\usepackage{times}
\usepackage{epsfig}
\usepackage{graphicx}
\usepackage{amsmath}
\usepackage{amssymb}
\usepackage{times}
\usepackage{epsfig}
\usepackage{graphicx}
\usepackage{amsmath}
\usepackage{amssymb}
\usepackage{xcolor}
\newcommand{\methodName}{{ChoiceNet}}
\newcommand{\method}{{Choice}}

\usepackage[breaklinks=true,bookmarks=false]{hyperref}
\wacvfinalcopy % *** Uncomment this line for the final submission

\def\wacvPaperID{614} % *** Enter the wacv Paper ID here

\ifwacvfinal\fi
\begin{document}

%%%%%%%%% TITLE

\title{\methodName: CNN learning through choice of multiple feature map representations}
\author{%Farshid Rayhan, \\
	%School of Computer Science  \\
	%Unversity of Manchester\\
	%{\tt\small farshid.rayhan@postgrad.manchester.ac.uk}
	Farshid Rayhan\textsuperscript{*}\\
	School of Computer Science\\
	The University Of Manchester \\
	{\tt\small farshid.rayhan@} \\ {\tt\small manchester.ac.uk}
	\and
	Aphrodite Galata\\
	School of Computer Science\\
	The University Of Manchester \\
	{\tt\small A.Galata@manchester.ac.uk 	}
	\and
	Timothy F. Cootes\\
	School of Health Sciences\\
	The University of Manchester\\
	{\tt\small 	t.cootes@manchester.ac.uk 	}
	%\and
	%Fourth Author\\
	%Department\\
	%school\\
	%email@edu
}
\maketitle
\ifwacvfinal\thispagestyle{empty}\fi

%%%%%%%%% ABSTRACT
\begin{abstract}
 		We introduce a new architecture called \methodName\ where each layer of the network is highly connected with skip connections and channelwise concatenations. This enables the network to alleviate the problem of vanishing gradients, reduces the number of parameters without sacrificing performance, and encourages feature reuse. We evaluate our proposed architecture on three benchmark datasets for object recognition tasks (ImageNet, CIFAR- 10, CIFAR-100, SVHN) and on a semantic segmentation dataset (CamVid).        		
\end{abstract}

%%%%%%%%% BODY TEXT
\section{Introduction}

{\color{black}Convolutional networks have become a dominant approach for visual object recognition \cite{he2016deep,simonyan2014very,lecun1989backpropagation,srivastava2015training}. However, as Convolutional Neural Networks (CNNs) are becoming increasingly deep, the vanishing gradient problem \cite{he2016deep} poses significant challenges as input information can vanish passing through many layers before reaching the end.}

% Recent research addressed this problem by either using skip connections \cite{he2016deep, srivastava2015training,larsson2016fractalnet} or by feature map concatenation \cite{huang2017densely}. 

{\color{black}When training a deep neural network gradients can become very small during the backpropagation process, making it hard to optimise the parameters in the early stages of the network. Therefore in the training phase the weights of the layers at the end of the network get updated quite rapidly while the early layers do not, leading to poor results. Activation function 'ReLU' and regularization methods like dropouts were proposed to address this problem \cite{dahl2013improving}. However, while these methods are important they do not solve the problem entirely. Huang et al. \cite{huang2016deep} found that as layers are added  to a network, at some point its performance will start to decrease \cite{huang2016deep}. Recent work \cite{he2016deep, huang2017densely,szegedy2015going,szegedy2017inception} proposed different solutions such as skip connections  \cite{he2016deep}, use of different sized filters in parallel  \cite{szegedy2015going,szegedy2017inception} and exhaustive concatenation between layers \cite{huang2017densely}. This goes some way to addressing the problem.    
}

\begin{figure}
	\centering
	\includegraphics[scale=.3]{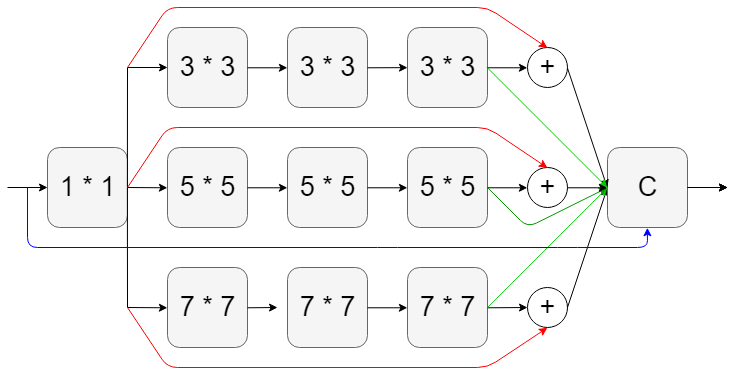}
	\caption{A single module of \methodName. The "+" denotes skip connections and "C" denotes channel wise concatenation.}
	\label{Model}
\end{figure}

In this paper we draw inspiration from the above networks \cite{he2016deep,huang2017densely} and propose a novel network architecture that retains positive aspects of these approaches \cite{he2016deep,huang2017densely}  whilst overcoming some of their limitations. Figure \ref{Model} illustrates a single module layout of our proposed architecture where its unique connectivity is displayed. 
We show that \methodName\ design allows good gradient and information flow through the network while using  fewer parameters compared to other state of the art schemes. We evaluate \methodName\ on three benchmark datasets (CIFAR10 \cite{krizhevsky2009learning}, CIFAR 100 \cite{krizhevsky2009learning} and SVHN \cite{netzer2011reading}) for image classification and also compare the performance of our network with state of the art methods in CamVid dataset \cite{kendall2015bayesian}. Our model performs well against existing networks \cite{he2016deep, huang2017densely} on all three datasets, showing promising results when compared to the current state-of-the-art.

\begin{figure}
	\centering
	\includegraphics[scale=.3]{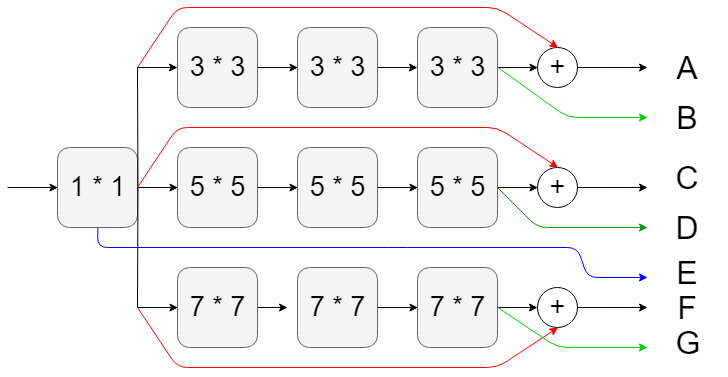}
	\caption{A breakdown of the \methodName\ module of Fig. \ref{Model}. Here letters A to G denote unique information generated by one forward pass through the module.}
	\label{Model6}
\end{figure}

\begin{figure*}
	\centering
	\includegraphics[scale=.2]{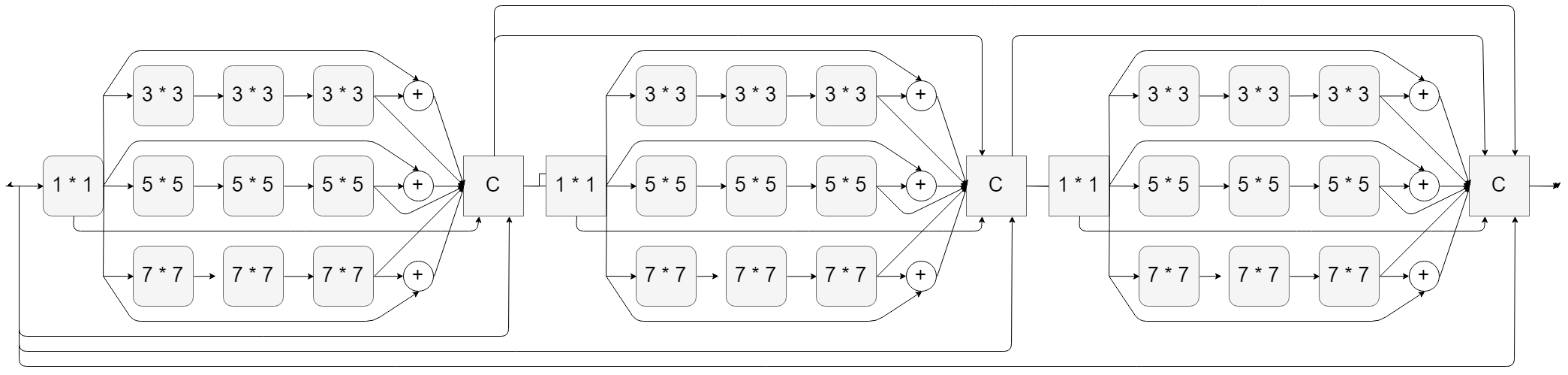}
	\caption{A single block of \methodName\ containing three consecutive \methodName\ modules (see fig: \ref{Model}). They are simply stacked one after another and densely connected like DenseNet \cite{huang2017densely}.}
	\label{Model-block}
\end{figure*}

\section{\methodName}

%Since the discovery of convolutional network, finding the ideal network architecture for a particular task has been a challenging area of research. The increased number of layers in modern architectures signifies the differences between different pattern of connectivity and revising older ideas.       

Consider a single image $x_0$ that is passing through a CNN. The network has $L$  layers, each with a non-linear transformation $H_l(.)$, where $l$ is the index number of the layer. $H_l(.)$ is a list of operations such as Batch-Normalization \cite{ioffe2015batch}, Pooling \cite{lecun1998gradient}, rectified linear units \cite{nair2010rectified} or a convolutional operation. The output of $l^{th}$ layer is denoted as $x_l$.

{\color{black}
	
	\textbf{ResNet}: ResNet \cite{he2016deep} uses identity mapping as bypassing paths to improve over a typical CNN network \cite{Krizhevsky2009}.

	A typical convolutional feed-forward network connects the $l^{th}$ layer's output to the $(l+1)^{th}$ layer's input. It gives rise to a layer transition: $x_l$ = $H_l(x_{l-1})$. ResNet \cite{he2016deep} adds an identity mapped connection, also referred as skip connection, that bypasses the transformation in between:
	
	%\begin{center}
	\begin{equation}
	x_l = H_l(x_{l-1}) + x_{l-1}
	\end{equation}      
	%	\label{resnet}
	%\end{center}
	
	This mechanism allows the network to flow gradients directly through the identity functions which results in faster training and better error propagation. However in \cite{huang2017densely} it was argued that despite the benefits of using skip connections, there is a possibility that when a layer is connected by a skip connection it may disrupt the information flow of the network therefore degrading the performance of the network.

	In \cite{targ2016resnet}, a wider version of ResNet was proposed where the authors showed that an increased number of filters in each layer could improve the overall performance with sufficient depth. FractalNet \cite{larsson2016fractalnet} also shows comparable improvement on similar benchmark datasets\cite{krizhevsky2009learning}.
	
}

\textbf{DenseNet}: As an alternative to ResNet, DenseNet proposed a different connectivity scheme. They allowed connections from each layer to all of its subsequent layers. Thus $l_{th}$ layer receives feature maps from all previous layers. Considering $x_0,x_1...,x_{l-1}$ as input:   

%\begin{center}
\begin{equation}
x_l = H_l([x_0,x_1...,x_{l-1}]) 
\end{equation}
%	\label{densenet}      
%\end{center}

where $[x_0,x_1...,x_{l-1}]$ denotes the concatenation of feature maps produced from previous layers respectively.

The network maximizes information flow by connecting the convolutional layers channel wise instead of skipping connections. In this model, the layer $l$ has $l$ number of inputs consisting of all the feature maps of previous $l-1$ layers. Thus on the $l_{th}$ layer, there are $l(l+1)/2$ connections. DenseNet requires fewer parameters as there is no need to learn from redundant features maps.  This allows the network to compete against ResNet using fewer parameters.

\textbf{\methodName}: We propose an alternative connectivity that retains the advantages of the above architectures whilst reducing some of their limitations. Figure \ref{Model} illustrates the connectivity layout between each layer of a single module. Each block of \methodName\ contains three modules and the total network is comprised of three blocks with pooling operations in the middle (see Figure \ref{Model-block}).

{\color{black}
	Figure \ref{Model6} shows a breakdown of each module. Letters A to G denote unique information generated by one forward pass through the model. B is generated by three consecutive $3 \times 3$ convolutional operations, whereas A is the result of the same three convolutional operations but additionally connected by a skip connection. Following this pattern, we generate information represented by letters C, D, F and G. Letter E denotes the special case where no convolutional operation is done after the $1 \times 1$ convolutional operation and it contains all the original information. This information is then concatenated with the others (ie. A, B etc.) at the final output.
	
	Therefore, the final output contains information with and without skip connections from filters of size 3, 5 and 7 and also from the original input without any modification. Note that the $1 \times 1$ convolutional operation at the start acts as a bottleneck to limit computational costs and all the convolutional operations are padded appropriately for the concatenation at the final stage. 
}

Considering $x_0, x_1..., x_{l-1}$ as input, our proposed connectivity is given by:

\begin{equation}
x_l = H_l(x_{l-1}) + x_{l-1} 
\end{equation}

\begin{equation}
x_{l+1} = H_l([x_{l},x_{l-1}] ) + x_{l}
\end{equation}

where $[x_l, x_{l-1}]$ is concatenation of feature maps. The feature maps are first summed and then concatenated which resembles characteristics of ResNet and DenseNet respectively. 

\textbf{Composite function}: Each of the composite functions consists of a convolution operation followed by a batch normalisation, and ends with a rectified linear unit(ReLU) operation. 

\begin{figure*}
	\centering
	\includegraphics[scale=.12]{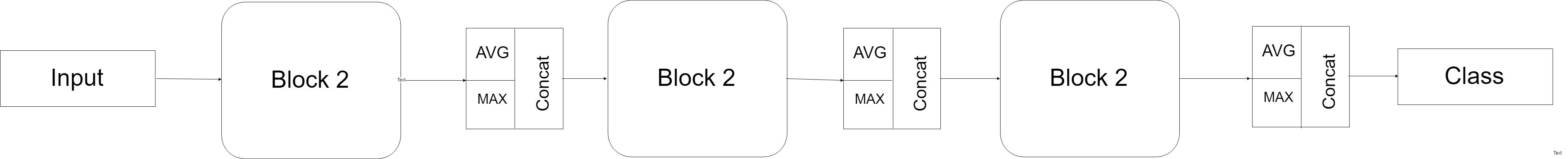}
	\caption{The \methodName\ consists of three \methodName\ blocks where each block contains three \methodName\ modules (see Fig. \ref{Model-block}) and each \methodName\ module is connected via feature maps and skip connections (see Fig. \ref{Model}). After each block, there is a Max-pool and an Avg-Pool operation and their feature maps are concatenated for the next layer.       }
	\label{Model-main}
\end{figure*}

\textbf{Pooling}: Pooling is an essential part of convolutional networks since Equations 1 and 2 are not viable when the feature maps are not of equal size. We divide the network into multiple blocks where each block contains same sized features. Instead of using either max pooling or average pooling, we use both pooling mechanisms and concatenate them before feeding it to the next layer (see figure \ref{Model-main}).       

\textbf{Bottleneck layers}: The use of $1\times1$ convolutional operations (known as bottleneck layers) can reduce computational complexity without hurting the overall performance of a network \cite{lin2013network}.  
%{\color{red}DenseNet has two alternative network implementations, one that includes bottleneck-layers and another that does not. Unlike DenseNet,}
We introduce a $1\times1$ convolutional operation at the start of each composite function (see fig \ref{Model} and \ref{Model-block}).

\textbf{Implementation Details}: \methodName\ has three blocks with equal number of modules inside. In each \method\ operation (see fig \ref{Model}), there are three $3\times3$, three $5\times5$ and three $7\times7$ convolutional operations. Each of the consecutive convolutional operations is connected via a skip connection (red line in fig \ref{Model}). The feature maps are then concatenated so that both the outputs with the skip and without the skip connections are included (green and black lines in fig \ref{Model} before "C"). Finally, the original input feature map is also merged (blue line in fig \ref{Model}) to produce the final output.

{\color{black}
	The intuition behind having the skip (Letter A, Figure \ref{Model6}) and the non-skip connections (Letter B, Figure \ref{Model6}) output merged  together is for enabling the network to choose between the two options for each filter size. We also merge the original input to this output (Letter E Figure \ref{Model6}) so that the network can choose a suitable depth for optimal performance. To allow the network further options, we use both Max and average pooling. Thus, each pooling layer contains both a Max-Pool and an Avg-Pool operation. The outputs of each pooling operation are merged before proceeding to the next layer. 
}
%The first two pooling operation had kernel size $= 2$. And the third pooling operation had kernel size $=$ 7 following a fully connected layer with $n$ neurons where $n$ denotes the number of classes.  

\section{Experiments}

We evaluate our proposed \methodName\ architecture on three benchmark datasets (CIFAR10 \cite{krizhevsky2009learning}, CIFAR 100 \cite{krizhevsky2009learning} and SVHN \cite{netzer2011reading}) and compared it with other state of the art architectures. We also evaluated it on state of the art semantic segmentation dataset CamVid \cite{kendall2015bayesian}.

\subsection{Datasets}
\subsubsection{CIFAR}
The CIFAR dataset \cite{krizhevsky2009learning} is a collection of two datasets, CIFAR10 and CIFAR100. Each dataset consists of 50,000 training images and 10,000 test images with $32\times 32$ pixels. The CIFAR10 dataset contains 10 class values and CIFAR100 dataset contains 100. In our experiment, we hold out 5,000 images from the training set for validation and use the rest of the images for training. We choose the model with the highest accuracy on the validation set to test on the testset. We adopt standard data augmentation with training including horizontally flipping images, random cropping, shifting and normalizing using channel mean and standard deviation. These augmentations were widely used in previous work \cite{he2016deep, huang2016deep,larsson2016fractalnet,lee2015deeply,lin2013network,romero2014fitnets,springenberg2014striving,srivastava2015training}. We also tested our model on the datasets without augmentation. In our final output in Table \ref{TabResult}, we denote the original dataset as C10 and C100, and the augmented dataset as C10+ and C100+.           

\subsubsection{SVHN}
The SVHN dataset contains images of \textbf{S}treet \textbf{V}iew \textbf{H}ouse \textbf{N}umbers with $32\times32$ pixels.There are 73,257 images in the training set and 26,032 on the testset. It also contains additional 531,131 images for training purposes. Like in previous work \cite{he2016deep, huang2016deep,larsson2016fractalnet,lee2015deeply,romero2014fitnets}, we use all the training data with no augmentation and use 10\% of the training images as a validation set. We select the model with the highest accuracy on the validation set and report the test error in Table \ref{TabResult}.        

%{\color{blue}
\subsection{ImageNet}
The ILSVRC  2012 classification dataset \cite{russakovsky2015imagenet} consists of 1.2 million images for training, and 50,000 for validation with 1, 000 classes. We adopt the same data augmentation scheme for training images as in \cite{huang2016deep,huang2017densely} and apply a single-crop or 10-crop with size 224 $\times$224 at test time. Following \cite{huang2017densely}, we report classification errors on the validation set.
%}

\subsubsection{CamVid}
The CamVid dataset \cite{fauqueur2007assisted} is a dataset consisting of 12 classes and has been mostly used for the task of semantic segmentation in previous work \cite{mulalic2018object,badrinarayanan2017segnet,cordts2016cityscapes}. The dataset contains a training set of 367 images, a validation set of 100 images and a test set of 233 images. The challenge is to do pixel wise classification of the input image and correctly identify the objects in the scene. The metric called IoU or 'intersection over union' is commonly used for this particular task \cite{chen2018deeplab,kendall2015bayesian,badrinarayanan2017segnet}.

\subsection{Training}
\subsubsection{Classification}

All networks were trained using stochastic gradient decent (SGD) \cite{bottou2010large}. We avoid using other optimizers such as Adam \cite{kingma2014adam} and RMSProp \cite{graves2013generating} to keep the comparisons as fair and simple as possible. On all three datasets, we used a training batch of 128. For the first 100 epochs, we used a learning rate of $0.001$, for the next 100 epochs $0.0001$, and then a rate of $0.00001$ for the final 300 epochs.

%{\color{blue} 
%Similar to DenseNet, \methodName\ contains some memory inefficiency due to its similar characteristics. From the technical report \cite{pleiss2017memory} those inefficiencies are discussed in details, it was found that there is a trade of between memory and accuracy. Thus the results provided on the DenseNet article could not be achieved if the efficient version was used.}
{\color{black}
	We use the learning parameters from \cite{huang2016deep} which was later used by \cite{huang2017densely} in order that the training environment is the same for every network.  
}

\subsubsection{Segmentation}
\begin{figure}
	\centering
	\includegraphics[scale=.4]{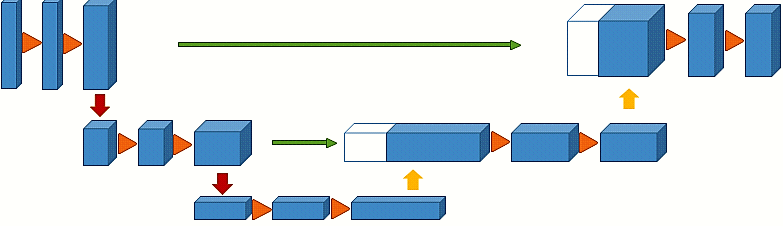}
	\caption{Training procedure using U-Net  \cite{ronneberger2015u}. Before each pooling operation, the features are stored and later concatenated when the feature maps are upsampled as indicated by the green arrows.  }
	\label{unet}
\end{figure}

For this task we use the training procedure of U-Net \cite{ronneberger2015u} (Fig. \ref{unet}) and we change the conv-blocks of U-Net with Res-Block (a block of the network that holds off the unique properties), Dense-Block and \methodName-Module (Fig. \ref{Model-block}). We use the Adam Optimizer with an initial learning rate of $0.001$ which was reduced by a factor of 10 after each 100 epochs until the network converged. A weight decay of $0.0005$ and Nesterov \cite{ruder2016overview} momentum without dampening was used. For fair comparison we kept the number of channels of Res-block and Dense-block unchanged as in the original article \cite{huang2017densely,huang2016deep}. 

Each of the experiments was performed 5 times and during the training process we took the model with the best validation score and reported its performance on the test set. 

%{\color{blue}

\subsubsection{Setup}
We used PyTorch \cite{paszke2017automatic} to implement our models. We used a machine with 16Gb of RAM with Intel i7 8700 with 6 core CPU and a Nvidia RTX 2080ti with 11GB of VRAM. 

%We trained DenseNet and ResNet models using Pytorch's implementations \cite{paszke2017automatic}. 
%DenseNet was initially programmed in language 'Lua' \cite{report1} and the authors later created a python version, but it was found that neither the author's version or the Pytorch's own implementation was able to reproduce the same result. It was also reported in \cite{report2} and \cite{report3} that with any Pytorch implementations the result are much lower ($0.5$ up to $1\%$ lower) for DenseNet and ResNet. However this is also considered quite common dues to usage of different language and library functions \cite{report2}. For this reason, 
%We ran all the experiments in Python language with Pytorch library to make the comparison as fair as possible with \methodName. 

%}

\begin{table*}[htb!]
	\begin{center}
		%\resizebox{\textwidth}{!}{%
		\scalebox{1}{
			%\begin{scriptsize}	
			\begin{tabular}{c|cc|ccccc} 
				\hline
				Method                       & Depth & Params & C10   & C10+ & C100  & C100+ & SVHN \\ \hline
				Network in Network           & -     & -      & 10.41 & 8.81 & 35.68 & -     & 2.35 \\ \hline
				All-CNN                      & -     & -      & 9.08  & 7.25 & -     & 33.71 & -    \\ \hline
				Deeply Supervised Net        & -     & -      & 9.69  & 7.97 & -     & 34.57 & 1.92 \\ \hline
				Highway Network              & -     & -      & -     & 7.72 & -     & 32.39 & -    \\ \hline
				FractalNet                   & 21    & 38.6M  & 10.18 & 5.22 & 35.34 & 23.3  & 2.01 \\
				with Dropout/Drop-path       & 21    & 38.6M  & 7.33  & 4.6  & 28.2  & 23.73 & 1.87 \\ \hline
				ResNet                       & 110   & 1.7M   & -     & 6.61*& -     & -     & -    \\
				ResNet (reported by \cite{huang2016deep} )       & 110   & 1.7M   & 13.63 & 6.41 & 44.74 & 27.22 & 2.01 \\ \hline
				ResNet with Stochastic Depth & 110   & 1.7M   & 11.66 & 5.23 & 37.8  & 24.58 & 1.75 \\
				& 1202  & 10.2M  & -     & 4.91 & -     & -     & -    \\ \hline
				Wide ResNet \cite{zagoruyko2016wide}                 & 16    & 11.0M  & 6.29 & 4.81 & -     & 22.07 & -    \\
				& 28    & 36.5M  & -     & 4.17 & -     & 20.5  & -    \\
				with Dropout                 & 16    & 2.7M   & -     & 4.2 & -     & -     & 1.63 \\ \hline
				ResNet (pre-activation)      & 164   & 1.7M   & 10.5* & 5.83 & 35.78 & 24.34 & -    \\
				& 1001  & 10.2M  & 10.4* & 4.59 & 32.89 & 22.75 & -    \\ \hline
				DenseNet (k = 12)            & 40    & 1.0M   &  7.0   & 5.24* & 27.55* & 24.42* & 1.79* \\
				DenseNet (k = 12)            & 100   & 7.0M   & \bf5.77*  & 4.1*  & 23.79* & 20.24* & 1.67* \\
				DenseNet (k = 24)            & 100   & 27.2M  & 5.83*  & \bf3.74*  & \bf23.42* & \bf19.25*  & \bf1.6* \\ \hline
				DenseNet-BC (k = 12)         & 100   & 0.8M   & 6*  & 4.51*  & 24.60*  & 22.98* & 1.76* \\
				DenseNet-BC (k = 24)         & 250   & 15.3M  & \bf5.16*  & 3.9*  & \bf19.75* & \bf17.60 & \bf1.74 \\
				DenseNet-BC (k = 40)         & 190   & 25.6M  & -     & \bf3.7*   &  -     & 17.88* & -    \\ \hline
				
				Inception v3 				 & 	-    & 13M   & 6.7    & 6.2   &  24.75	  & 23.5  & 1.8  \\
				Inception v4 				 & -     & 18M   & 6.4    & 6.0   &  24.40	  & 23.22 & 1.75 \\ \hline
				\methodName-30  & 30  	& 13M  	  & 5.9  & 4.2 &  22.80		& 20.5 & 1.8 \\
				\methodName-37  & 37   & 19.2M   & \color{blue}4.0  & \color{blue}3.9 & \color{blue} 18.91 & \color{blue}17 & \color{blue}1.6 \\
				\methodName-40  &  40  & 23.4M  	  & \color{blue}3.9 & \color{blue}3.2 & \color{blue}18.05  & \color{blue}16.6 & \color{blue}1.5  \\ \hline			
			\end{tabular}%
		}
		
		%\end{scriptsize}
	\end{center}
	\caption{Error rates ( 100 - accuracy )\% on CIFAR and SVHN datasets. $k$ denotes network’s growth rate for DenseNet. Results that surpass all competing methods are blue and the overall best results are bold. “+” indicates standard data augmentation (translation and/or mirroring). The notation '*' indicates models run by ourselves. All the results of \methodName\ without data augmentation(C10, C100, SVHN) are obtained using Dropout. \methodName\ achieve lower error rates while using fewer parameters than ResNet and DenseNet. Without data augmentation, \methodName\ performs better by a significant margin.
	}
	\label{TabResult}
\end{table*}

\subsection{Result Analysis}
\subsubsection{CIFAR and SVHN}

\textbf{Accuracy}:  Table \ref{TabResult} shows that the \methodName\ depth 40 achieves the highest accuracy on all three datasets. The error rate on C10+ and C100+ is 4.0\% and 17.5\% respectively which is lower than error rates achieved by other state of the art models. Our results on the original C10 and C100 (without augmentation) data sets are  2\% lower than Wide ResNet and 5\% lower than pre-activated ResNet. Our model \methodName\ ($d = 37$) performs comparably well to DenseNet-BC with $k = 24$ and $k = 40$, whereas \methodName\ ($d=40$) outperforms all other networks.                

\textbf{Parameter efficiency}: Table \ref{TabResult} shows that \methodName\ needs fewer parameters to give similar or better performance compared  to other state of the art architectures. For instance, \methodName\ with a depth of 30 has only 13 million parameters yet it performs comparably well to DenseNet-BC ($k=24$) which has 15.3 million parameters. Our best results were achieved by \methodName\ ( $d = 40$ ) with 23.4 million parameters compared to DenseNet-BC ( $k = 40$ ) with 25.6m, DenseNet ( $k = 24$ ) with 27.2m and Wide ResNet with 36.5m parameters.           

\textbf{Over-fitting}:
Deep learning architectures can often be prone to overfitting however as \methodName\ requires a smaller number of parameters, it is less likely to overfit the training datasets. Its performance on the non-augmented  datasets appears to support this claim.         

\textbf{Exploding Gradient}:
While training \methodName\ we observed that it occasionally suffers from an exploding gradient problem. ResNet and DenseNet were both trained using stochastic gradient descend(SGD) and a learning rate of $0.1$ that was later reduced to $0.01$ and $0.001$ after every 100 epochs. However, we had to start training our network using a learning rate of  $0.001$ because setting the rate any higher was causing gradients to explode. We also had to reduce the learning rate to $0.0001$ and then to $0.00001$ after each 50 epochs instead of 100 to prevent the problem from reoccurring. 

The problem of exploding gradients is easier to handle than that of vanishing gradients. We used a smaller learning rate at the start and L2 regularisers with dropout layers ($p = 0.5$) which addressed the problem.          

\begin{figure}
	\centering
	\includegraphics[scale=.58]{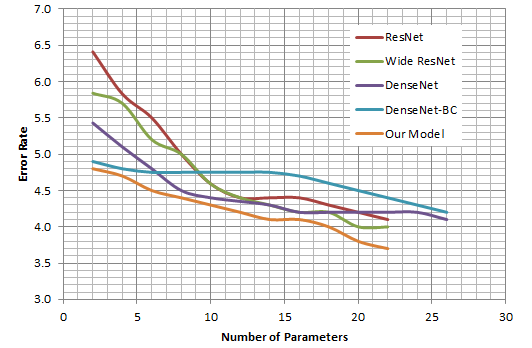}
	\caption{Error vs parameters (in millions) comparison between ResNet, Wide ResNet, DenseNet, DenseNet-BC and \methodName\ on C10+ dataset. Each of the models were trained on same environment with different depths and number of parameters and later they were plotted with a smoothed line for improved clarity.}
	\label{Compare}
\end{figure}

\begin{table}[htb!]
	\begin{center}
		%\resizebox{\textwidth}{!}{%
		\scalebox{1}{
			%\begin{scriptsize}	
			\begin{tabular}{c|cc} 
				\hline
				Method                			& Top-1 Error\(\%\) & Top-5 Error\(\%\) \\ \hline
				
				Inception v3 \cite{szegedy2016rethinking} & 22.9 & 	5.9   \\
				Inception v4 \cite{szegedy2017inception}  &  21.5 & \bf5.7  \\ 		
				
				ResNet-50    &  22.85 &6.71  \\ 
				ResNet-101   &  21.75 & 6.05  \\ 
				ResNet-152   &  \bf21.43 & 5.71  \\ 
				
				DenseNet-121  &  25.02 & 7.71\\
				DenseNet-169  &  23.82 & 6.85\\
				DenseNet-201  &  22.58 & 6.34\\
				DenseNet-264  &  22.15 & 6.12\\  \hline
				
				\methodName-30    &   \color{blue}21.30 & 5.8    \\
				\methodName-37   &   \color{blue}21.21  & \bf5.7    \\
				\methodName-40   &  \color{blue}20.53 &  \color{blue} 5.5    \\

			\end{tabular}
		}
		
		%\end{scriptsize}
	\end{center}
	\caption{Error rates of Top1 and Top 5\% on ImageNet dataset. Results that surpass all competing methods are blue and the overall best results are bold. \methodName\ achieves lower error top 1\% error rates with all three versions and lower top 5\% error with \methodName-40. }
	\label{TabResult3}
\end{table}

\begin{table}[htb!]
	\begin{center}
		%\resizebox{\textwidth}{!}{%
		\scalebox{1}{
			%\begin{scriptsize}	
			\begin{tabular}{c|c|cc} 
				\hline
				Method               & Pooling 		& Top-1 Error\(\%\) & Top-5 Error\(\%\) \\ \hline

				\methodName-30   & Max &  21.45 & 6.0    \\
				\methodName-37   & Max &  21.32 & \bf5.9    \\
				\methodName-40   & Max &  \bf21.02 & \bf5.9    \\ \hline
				
				\methodName-30   & Avg &  22.25 & 6.2    \\
				\methodName-37   & Avg &  22.21  & 6.1    \\
				\methodName-40   & Avg & 21.80 & \bf5.9    \\  \hline

				\methodName-30   & Both & 21.30 & \color{blue}5.8    \\
				\methodName-37   & Both & 21.21  &\color{blue}5.7    \\
				\methodName-40   & Both & \color{blue}20.53 & \color{blue} 5.5    \\

			\end{tabular}
		}
		
		%\end{scriptsize}
	\end{center}
	\caption{Error rates of Top1 and Top 5\% on ImageNet dataset of \methodName with only Maxpool, AvgPool and both of them together. Results that surpass all competing methods are blue and the overall best results are bold. \methodName\ achieves lower error top 1\% error rates with all three versions and lower top 5\% error with \methodName-40. }
	\label{TabResult4}
\end{table}

\begin{table}[htb!]
	\begin{center}
		%\resizebox{\textwidth}{!}{%
		\scalebox{1}{
			%\begin{scriptsize}	
			\begin{tabular}{c|cc} 
				\hline
				Method                						& m\_IoU \\ \hline
				%		Wu et. al. \cite{wu2019wider}         							 & 80.6       \\
				%		Wang et. al. \cite{wang2018understanding}				 		 & 80.1     \\
				%		Ke et. al. \cite{ke2018adaptive}								 &79.1 \\
				%		Kong et. al. \cite{kong2018recurrent}					 		& 78.2 \\
				%		Wang et. al. \cite{wang2017understanding} & 77.6 \\
				\methodName-block    &  \bf73.5   \\
				
				Inception v3 \cite{szegedy2016rethinking} & 71.9 	   \\
				Inception v4 \cite{szegedy2017inception}  &  71.5 \\ 		
				Lin et. al. \cite{lin2017refinenet} &73.6 \\
				Res-Blocks    &  70.6  \\ 
				
				%		Chen et. al. \cite{chen2018deeplab} & 70.4 \\
				%		Mehta et. al. \cite{mehta2018espnetv2} & 70.2 \\
				
				%		Fourure et. al. \cite{fourure2017residual} & 69.8 \\
				Dense-blocks   &  69.2\\
				
				Lo et. al. \cite{lo2018efficient} & 67.3 \\
				Yu et. al. \cite{yu2015multi} & 67.1 \\
				Kreo et. al. \cite{kreso16gcpr} &66.3 \\
				
				Chen et. al. \cite{chen14semantic} &63.1 \\
				Berman et. al. \cite{berman2018lovasz} &63.1 \\
				Arnab et. al. \cite{higherordercrf_ECCV2016} &62.5 \\
				
				Huang et. al.\cite{mehta2018espnetv2} & 60.3\\
				
			\end{tabular}
		}
		
		%\end{scriptsize}
	\end{center}
	\caption{The mean IoU (m\_IoU) of all the classes on the CamVid dataset (test-set) where mean-IoU means the mean of IoUs of all the 12 classes. 	}
	\label{TabResult2}
\end{table}

\begin{figure*}
	\centering 
	\scalebox{.8}{
		
	\begin{tabular}{cccc}
		
		Ground Truth & DenseNet & ResNet & \methodName\  \\
		
		%	\includegraphics [width=0.22\textwidth]{g1.png}& \includegraphics [width=0.22\textwidth]{d1.png}&	\includegraphics [width=0.22\textwidth]{r1.png}
		%	& \includegraphics [width=0.22\textwidth]{o1.png} \\
		%		(a) & (b) & (c) & (d) \\
		
		\includegraphics [width=0.22\textwidth]{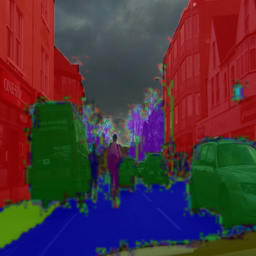}& \includegraphics [width=0.22\textwidth]{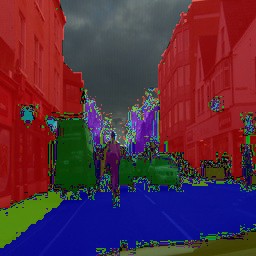}&	\includegraphics [width=0.22\textwidth]{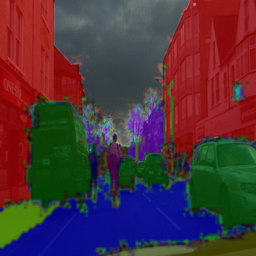}
		& \includegraphics [width=0.22\textwidth]{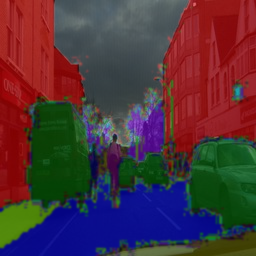} \\
		%		(a) & (b) & (c) & (d) \\		
		
		\includegraphics [width=0.22\textwidth]{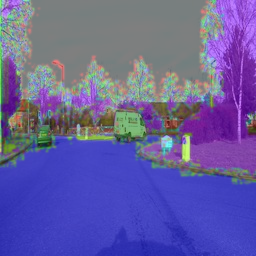}& \includegraphics [width=0.22\textwidth]{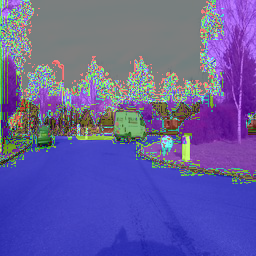}&	\includegraphics [width=0.22\textwidth]{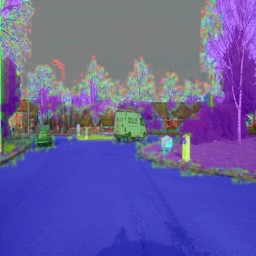}
		& \includegraphics [width=0.22\textwidth]{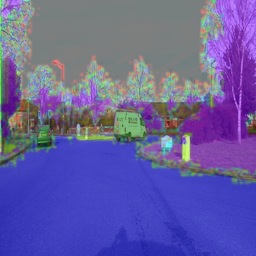} \\
		
		\includegraphics [width=0.22\textwidth]{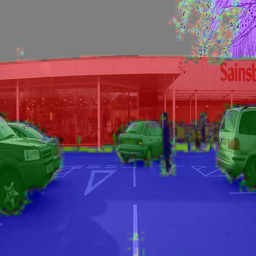}& \includegraphics [width=0.22\textwidth]{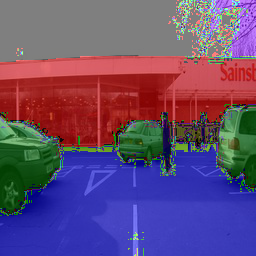}&	\includegraphics [width=0.22\textwidth]{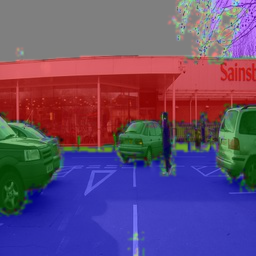}
		& \includegraphics [width=0.22\textwidth]{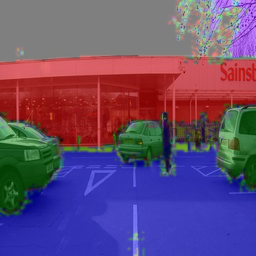} \\

	\end{tabular}
    }
	
	\caption{ Predicted output from ResNet, DenseNet and \methodName. Our Model displays superior performance, particularly with its ability to detect boundaries more precisely compared to a deeper ResNet model and a wider DenseNet model. Note that these results correspond to the model giving the highest mIoU accuracy}
	
	\label{segImages}

\end{figure*}

\subsubsection{CamVid}

We tested \methodName\ on the CamVid dataset and compared it with other state of the art networks \cite{wu2019wider,wang2018understanding,ke2018adaptive,kong2018recurrent,wang2017understanding,lin2017refinenet,chen2018deeplab,chen2018deeplab,mehta2018espnetv2,fourure2017residual}. Mean IoU (m\_IoU) scores are shown in table \ref{TabResult2}. 

Although our network performs better than other neural network architectures, it was able to outperform  DenseNet and ResNet both in terms of m\_IoU score as well as in terms of parameter efficiency. Our \methodName\ with 13 million parameters was able to perform better than networks almost twice its size. 

In Figure \ref{segImages} we display some of the predicted images from ResNet, DenseNet and our model against ground truth data from CamVid dataset. The qualitative results show that our model has the ability to segment smaller classes with good precision.

%\subsection{Cityscapes}  
%We tested \methodName\ on the Cityscapes dataset and compared the result with other state of the art networks \cite{wu2019wider,wang2018understanding,ke2018adaptive,kong2018recurrent,wang2017understanding,lin2017refinenet,chen2018deeplab,chen2018deeplab,mehta2018espnetv2,fourure2017residual}. Mean IoU (m\_IoU) scores are shown in table \ref{TabResult2}. 

%Although our network did not perform better than specialist state of the art segmentation networks, it was able to outperform  DenseNet and ResNet both in terms of m\_IoU score as well as in terms of parameter efficiency. Our \methodName\ with 13 million parameters was able to perform better than networks almost twice its size.

%In Figure \ref{segImages} we display some of the predicted images from ResNet, DenseNet and our model against ground truth data from CamVid dataset. The qualitative results show that our model has the ability to segment smaller classes with good precision.  

\section{Discussion}

%\methodName\ has quite similar characteristics of both DenseNet and ResNet aside for some minor changes (see Eq 3 \& 4) which includes both skip connections and feature maps concatenation of previous layers.    

\textbf{Model compactness}: As a result of the use of different filter sizes with feature concatenation and skip connections at every stage, feature maps learned by any layer in a block can be accessed by all subsequent layers. This extensive feature reuse throughout the network leads to a compact model.  

In figure \ref{Compare}, we showed a layer vs test error graph which demonstrates the compactness of \methodName\ with respect to other state of the art architectures. Note that for training different networks we kept the environment same but changed the depth and later smoothed the curve for better visualisation.  \methodName's curve always stays at the very bottom which means better error rate with fewer parameters/layers.                            

\textbf{Feature Reuse}: \methodName\ uses different filter sizes with skip connections and channel concatenation in each module (see fig \ref{Model}). In order to have a deeper and visual understanding of its operation, we took the weights of the first block (in \methodName-37) and normalized them to the range $[0,1]$. After normalizing the weights we mapped them to two sections, weights under 0.4 as white and over 0.4 as colored - see table \ref{explain}. We assumed that the weights less than 0.4 will have insignificant effect on the total performance. 
The figure shows that after the very first $1\times1$ convolution operation on the raw input, the conv operations with channel size 7 has more effect than size 3 and 5. In the second module all the conv operations' weights were under 0.4 which suggests that the model used either the feature maps of the earlier output by concatenation (red line between filter 5 and 7 of the middle module) or it used the skip connection (red line above filter 3 with highlighted "+" sign). On one hand this indicates that the skip connection or channel concatenation or both are working as they were suppose to but this also means that we still have many redundant parameters in the network. In the third module it was found that filters 3 and 5 had weights over 0.4 which indicates that they possibly had some contribution in the network. We suspect that the selection of filter size 7 in the first module and 3 and 5 in the third module echoes the hypothesis of AlexNet \cite{krizhevsky2012imagenet} where they found bigger filter sizes work better at the beginning of the networks and smaller filters work better in the later stages.   

In table \ref{TabResult2}, we show the \textit{\textbf{M}ean \textbf{I}ntersection \textbf{o}ver \textbf{U}nion} (m\_IoU) on the CamVid dataset of some of the current state of the art models. We used the U-Net training scheme and changed the basic convolutional operations with ResBlocks, DenseBlocks and \methodName-module (see figure \ref{Model}). While our network has fewer parameters compared to ResBlock and Denseblocks, it achieved a higher score. Note that even though our model achieved a good m\_IoU score , it is not as good as some of the network architectures designed specificaly for segmentation tasks \cite{wu2019wider,wang2018understanding,ke2018adaptive,kong2018recurrent,wang2017understanding}. Nevertheless, it performed  well comparing to both ResBlock and Dense-block as well as some other general purpose convolutional neural  networks \cite{mehta2018espnetv2}.

Our intuition is that 
%due to the use of skip connection and non-skip connection paths with different filter sizes and added the original input and concatenation all of them  at the end together (see fig \ref{Model}) 
the extra connections and paths in our method enable the network to learn from a large variety of feature maps. This also enables the network to back propagate errors more efficiently (see also \cite{he2016deep,huang2017densely}). We found that due to all the connections the network can be prone to exploding gradient and therefore needs a small learning rate to begin with. We also found by grid search that the network shows peak performance when the depth is between 30 to 40 layers and further increasing the layers appears to have little effect. We suspect that \methodName\ plateaus at depth 30 to 40 although it is possible that it could be a local minima as we couldn't train models with depth more than 60 layers due to resource limitation.   

{\color{black}
	The performance on ImageNet dataset is displayed in table \ref{TabResult3}. Our model with all three variation achieves lower top 1\% score compared to other state of the art neural network architectures like ResNet, DenseNet, Inception (v3/v4) and \methodName-40 scores the lowest top 5\% and top 1\% error. This is a result of the unique connectivity design (see figure \ref{Model6}). Due to the usage of convolutional output with and without skip connection, using different kernel sizes, concatenating the original input per module via the connection 'E' of figure \ref{Model6} and using two different pooling techniques together, it achieves this superior performance. Also as the architecture has many connections, therefore it can work with less channel outputs per convolution operation which makes it parameter efficient. This means given a number of parameters it achives better performance than other methods.         
	
}

{\color{black}
	In table \ref{TabResult4}, we show the effect of the usage of two types of pooling method with our architectural design. We find that for all three models the use of max pool gave advantage over avg pooling. \methodName-40 achieved the lowest error rate among the pooling techniques individually however it was superseded by the same model when both pooling were used. This shows even though, in cases, avg pooling may not be as effective as maxpool, using them together leads to  improved performance.

}

\begin{table*}[!htb]
	
	\begin{center}
		\begin{scriptsize}
			\begin{tabular}{l}
				
				\includegraphics[scale=.2]{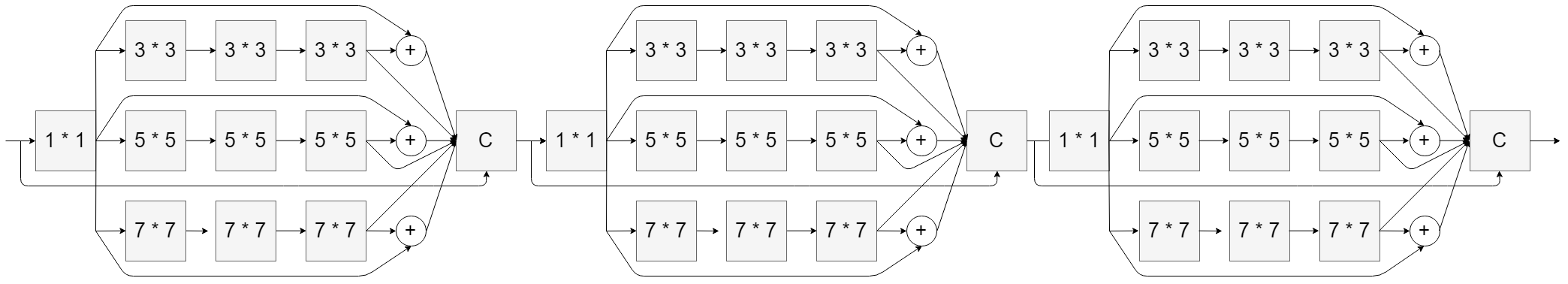} 	\\
				\includegraphics[scale=.2]{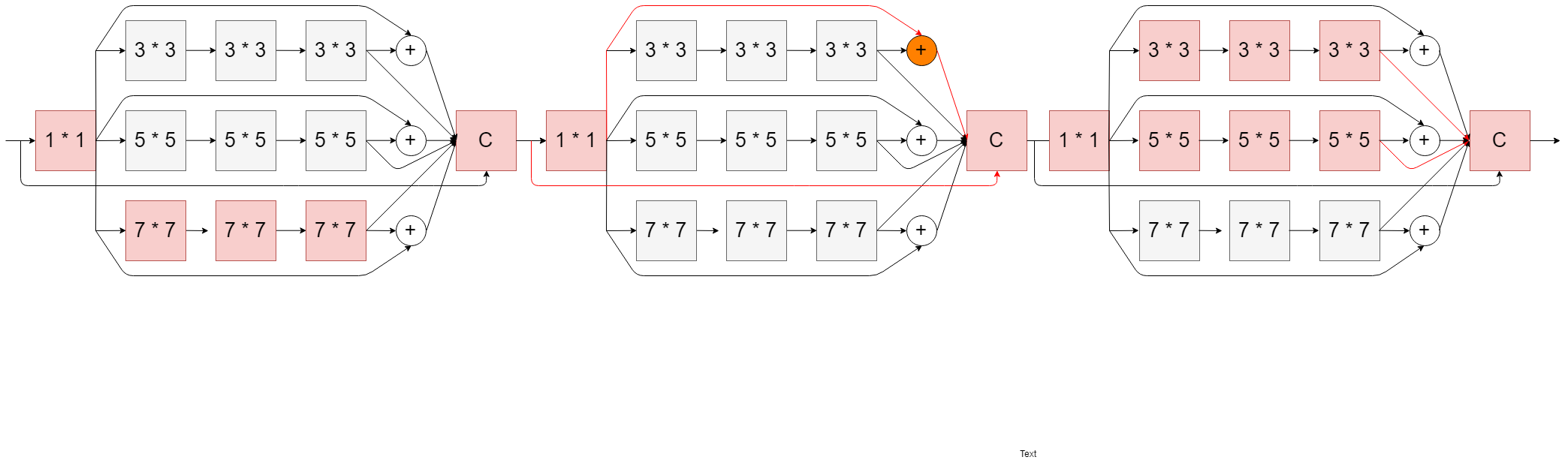}  \\
				
			\end{tabular}
		\end{scriptsize}
	\end{center}
	\caption{An inside look at \methodName. The top figure shows the skeleton of \methodName\ before training and the bottom figure shows a path way the model has chosen for C10* dataset for best classification accuracy after training. The colored boxes and lines are putting the most contribution.}
	\label{explain}
\end{table*}

\section{Conclusion}

In this paper, we introduced a powerful yet lightweight and efficient network, \methodName, which encodes better spatial information from images by learning from its numerous elements such as skip connections, the use of different filter size, dense connectivity and including both Max and Avg pooling. \methodName is a general purpose network
with good generalisation abilities and can be used across a
wide range of tasks including classification, image segmentation and others. Our network shows promising performance when compared to state-of-the-art techniques across different
tasks such as semantic segmentation and object classification while being more efficient.

{\small
\bibliographystyle{ieee}
\bibliography{egbib}
}

\end{document}